\documentclass[runningheads]{llncs}
\usepackage[T1]{fontenc}
\usepackage{hyperref}
\usepackage{url}
\usepackage{multirow}
\usepackage{multicol}
\usepackage{booktabs}       
\usepackage{graphicx}
\usepackage{subcaption}

\usepackage{xcolor}
\usepackage{algorithmicx}
\usepackage[ruled]{algorithm}
\usepackage{algpseudocode}
\usepackage{amsmath}

\def\our{MonAcoSFL}

\begin{document}
\title{
A deep cut into Split Federated Self-supervised Learning
}
\titlerunning{A deep cut into Split Federated Self-supervised Learning}
%

\author{
Marcin Przewięźlikowski\inst{1,2,3}\orcidID{0000-0003-4772-3268} \and
Marcin Osial\inst{1,2,3}\orcidID{0000-0002-7414-3608} \and
Bartosz Zieliński\inst{1,3}\orcidID{0000-0002-3063-3621} \and
Marek Śmieja\inst{1}\orcidID{0000-0003-2027-4132}
}
%
%
\institute{
Jagiellonian University, Faculty of Mathematics and Computer Science \and
Jagiellonian University, Doctoral School of Exact and Natural Sciences \and
IDEAS NCBR 
}
\maketitle              
\begin{abstract}
Collaborative self-supervised learning has recently become feasible in highly distributed environments by dividing the network layers between client devices and a central server. However, state-of-the-art methods, such as MocoSFL, are optimized for network division at the initial layers, which decreases the protection of the client data and increases communication overhead.
In this paper, we demonstrate that splitting depth is crucial for maintaining privacy and communication efficiency in distributed training. We also show that MocoSFL suffers from a catastrophic quality deterioration for the minimal communication overhead.
As a remedy, we introduce \textbf{Mo}me\textbf{n}tum-\textbf{A}ligned \textbf{co}ntrastive \textbf{S}plit \textbf{F}ederated \textbf{L}earning (\our{}), which aligns online and momentum client models during training procedure. Consequently, we achieve state-of-the-art accuracy while significantly reducing the communication overhead, making \our{} more practical in real-world scenarios\footnote{Our codebase is available at \href{https://github.com/gmum/MonAcoSFL}{\url{https://github.com/gmum/MonAcoSFL}}}.

\end{abstract}

\section{Introduction}

Collaborative learning techniques allow multiple participating parties to jointly train models without compromising the confidentiality of their private data, making them increasingly popular~\cite{mcmahan2017communication,thapa2020splitfed,zhang2021federated_survey,zhuang2022divergence,li2023mocosfl}. Among these techniques, Federated Learning (FL)~\cite{mcmahan2017communication} stands out as the most prevalent framework.
In FL, the learning task is solved by a federation of participating devices (which we refer to as clients) coordinated by a central server. Each client has a local training dataset (unseen by the server) and computes an update to the current global model maintained by the server (only this update is communicated). Moreover, a part of the model can be trained on the server side, resulting in a setup called Split Federated Learning (SFL)~\cite{thapa2020splitfed}.

Federated learning has exhibited considerable success in supervised learning tasks~\cite{mcmahan2017communication,chen2022vertical,wu2022federated}. However, the assumption of full labeling may sometimes be impractical due to the challenges and expertise required for accurate labeling~\cite{makhija2023federated}. That is why more practical methods focus on unlabeled data~\cite{zhuang2021collaborative,zhuang2022divergence}, combining FL with classic self-supervised learning (SSL)~\cite{grill2020bootstrap}. A notable example is MocoSFL, which is based on Split Federated Learning (SFL) and Momentum Contrast (MoCo)~\cite{he2020momentum,li2023mocosfl}.

\begin{figure}[t]
    \centering
    \includegraphics[width=0.75\textwidth]{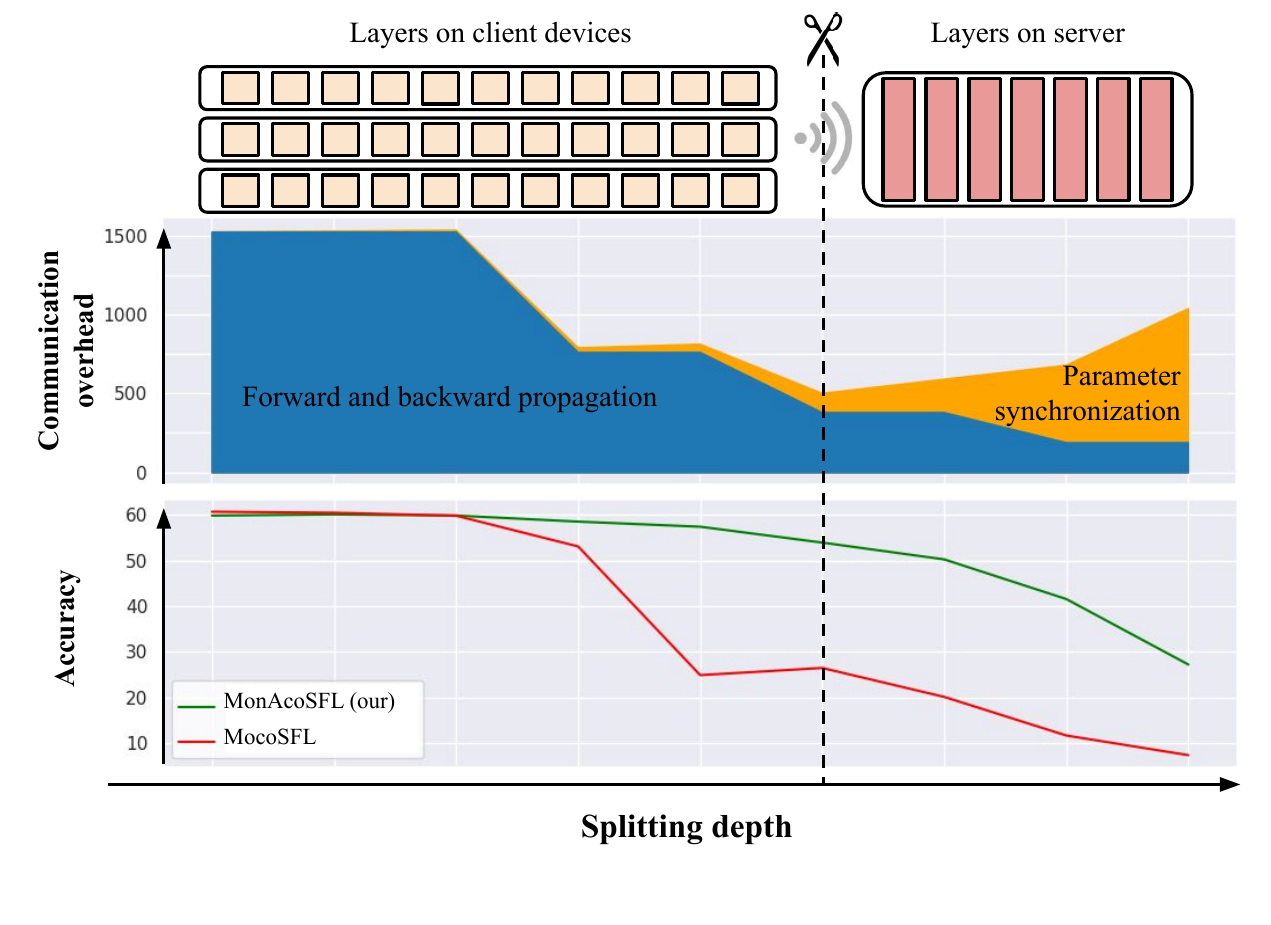}
    \caption{Communication overhead and accuracy for MocoSFL and \our{} depending on the splitting depth. In this example, the minimal communication overhead is obtained for dividing the model into 11 and 7 layers on the client and server sides, respectively. For such an optimal case, the accuracy of MocoSFL drops significantly, in contrast to the accuracy of \our{}. Notice that computational overhead consists of forward and backward propagation (marked as blue) and parameter synchronization (marked as orange).}
    \label{fig:teaser}
\end{figure}

MocoSFL can achieve good accuracy with relatively low memory requirements and can support a large number of clients. However, it was optimized for network division at the initial layers (one or three layers on client devices), which has certain disadvantages. First, it decreases the protection of the client data (Figure~\ref{fig:activations_privacy}). Secondly, it increases communication overhead (information transferred between clients and server), as shown in Figure~\ref{fig:teaser}.

In this paper, we delve into the relationship between communication overhead and the splitting point. We identify the optimal splitting point and highlight the poor performance of MocoSFL when aligned with it. As a remedy, we introducing \textbf{Mo}me\textbf{n}tum-\textbf{A}ligned \textbf{Co}ntrastive \textbf{SFL} (\our{}). In contrast to MocoSFL, which synchronizes only the online client models during the training, \our{} also synchronizes their momentum models. This change is crucial because it prevents the divergence of online and momentum models and reduces confusion during training (see Figure~\ref{fig:schema_misalignement}).

Through extensive experiments, we empirically confirm that the synchronizing scheme of MocoSFL indeed creates misalignment between its online and momentum models, which is not the case in \our{}. Crucially, we show that when training mobile-friendly architectures split at deeper, more communication-efficient points, \our{} improves over the performance of MocoSFL~\cite{li2023mocosfl} by a significant margin, confirming its practicality.  

\textbf{Our contributions can be summarized as follows:}
\begin{itemize}
    \item We analyze the practical trade-off in communication efficiency of split federated learning and show that MocoSFL performs poorly for the most optimal splits.
    \item We re-think model synchronization in MocoSFL and introduce \our{}, which reduces the divergence between online and momentum clients at each synchronization step.
    \item We present an extensive experimental study and confirm that \our{} maintains high performance even when applied to deeper splits of the backbone model, increasing its practical applicability to real-world scenarios.
\end{itemize}

\section{Related work}


\textit{Federated learning (FL)}~\cite{mcmahan2017communication,zhao2018federated,liu2021feddg,li2021fedbn} is a method of training models across multiple decentralized entities or devices while keeping the data localized, without the need to send it to a centralized location~\cite{mcmahan2017communication}. 
Models are individually trained and then aggregated by a global server, ensuring data privacy by sharing only model parameters among clients. 
The seminal FedAVG algorithm~\cite{mcmahan2017communication}, which conducts a straightforward weighted averaging of clients' model weights, remains a solid foundational framework for numerous approaches~\cite{zhuang2021collaborative,zhuang2022divergence,lee2023fedlp}. One of the most important variants of FL is Split Federated Learning (SFL), which involves storing a part of the trained model on a centralized server, reducing the computational overhead of clients at the cost of additional communication overhead due to sending latent vectors from clients to the server ~\cite{thapa2020splitfed}. 
In addition, SFL creates a trade-off between the clients' energy consumption and privacy~\cite{lee2024exploring}.


\textit{Self-supervised learning (SSL)}, a pivotal development in unsupervised representation learning, shines particularly in its joint-embedding form within the realm of computer vision \cite{caron2020unsupervised,caron2021emerging,he2020momentum,balestriero2023cookbook}. This approach leverages contrastive learning  \cite{bachman2019learning,oord2018representation} to modulate similarity and disparity among augmented views. To prevent trivial solutions, such methods often rely on substantial memory banks, like MoCo~\cite{he2020momentum,chen2020improved}, or necessitate large batch sizes as observed with SimCLR~\cite{chen2020simple}, to amass a significant number of meaningful negative examples. On the other hand, non-contrastive methods such as BYOL~\cite{grill2020bootstrap} do not require negative examples in their objectives, yet still require large batch sizes during training~\cite{grill2020bootstrap,chen2021exploring,caron2021emerging}. A common architectural component of several popular SSL approaches is the maintaining of "momentum" models to achieve asymmetry of representations~\cite{he2020momentum,chen2020improved,chen2021empirical,caron2021emerging,assran2023ijepa}.



Early attempts at deploying the SSL methods in Federated setting have 
grappled with the complexity of data diversity \cite{van2020towards} and addressed key privacy considerations \cite{zhang2023federated}. Methods like FedU \cite{zhuang2021collaborative} utilize the BYOL framework for federated SSL, while FedEMA \cite{zhuang2022divergence} refines this by adaptively updating client networks with insights from the global model's trends. However, the above methods cannot be scaled beyond cross-silo settings, involving up to 100 client devices. Li et.al. demonstrate the key ingredients of scaling SSL to highly distributed cross-client environments (as many as 1000 clients) in the form of MocoSFL -- a combination of MoCo and the Split Federated Learning paradigm~\cite{li2023mocosfl}.



\section{Preliminaries}

In this section, we first describe the problem of federated learning in a self-supervised regime and its challenges. Then, we recall the MocoSFL~\cite{li2023mocosfl}, a recently published method, which we use as a basis for our \our{}.

\subsection{Federated self-supervised learning}
\label{sec:ssl_and_federated}

\paragraph{Problem statement.}
In this paper, we tackle the problem of \emph{federated self-supervised learning}. It considers multiple participating devices (called clients) and a centralized server that collaboratively trains a single model based on data gathered by the former. However, the devices are prohibited from sending their raw data to the server e.g. for privacy reasons. Instead, the training is conducted in a \emph{federated learning} manner, where each client trains a copy of the model on its local data and periodically synchronizes it with other clients. Another assumption is that data gathered by the devices do not contain labels. Thus, the model must be trained by optimizing a \emph{self-supervised} objective. After the training, the resulting model can be finetuned to the target tasks using labeled data and distributed to all participating devices.

\paragraph{Split federated learning} (SFL)~\cite{mcmahan2017communication} is a practical approach to federated learning, which splits the model into client and server parts. For a formal definition, 
let us define $N$ as the number of participating clients and $f_{\phi_i}$ as a copy of model $f_\phi$ stored by the $i$-th client parametrized by $\phi_i$. Furthermore, let $f_{\phi}$ be composed of two parts $f^s_{\phi^s} \circ f^c_{\phi^c}$, where $\phi = \phi^s \cup \phi^c$, $f^c_{\phi^c}$ is distributed among client devices, and $f^s_{\phi^s}$ is stored on a centralized server. There are multiple copies of $f^c_{\phi^c}$ but only one version of parameters\footnote{Note that FL is a special case of SFL where $f = f^c$, whereas a situation where $f = f^s$ is equivalent to regular training on a single device.} $\phi^s$.

Client models start with the same initial parameters ($\phi^c_1 = \phi^c_2 = \dots = \phi^c_N$) and are trained in two phases:
\begin{enumerate}
    \item Optimization of $\phi^c_1, ..., \phi^c_N, \phi^s$ w.r.t. to training objective $\mathcal{L}$.
    \item Synchronization of $\phi^c_1, ..., \phi^c_N$ by overwriting each $\phi^c_i$ with $\hat{\phi^c} = \sum_{i=0}^N \dfrac{\phi^c_i}{N}$.
\end{enumerate}
Each client processes only its local data during the optimization step, which leads to increasingly diverging $\phi^c_1, ..., \phi^c_N$. That is why the synchronization step is required.

In standard Federated learning, the communication is limited to transferring parameters for model synchronization. In SFL, we must also transfer from clients the activations of $f^c$ (for forward propagation by the server) and their gradients w.r.t. to $\mathcal{L}$ from the $f^s$ residing on the server (for backpropagation in local models).

\paragraph{Self-supervised learning} (SSL) is a paradigm of learning representations without data labels~\cite{saleh2022selfsupervisedsurvey,balestriero2023cookbook}. Currently, the prevalent approaches to SSL are the joint-embedding architectures~\cite{he2020momentum,chen2020improved,chen2020simple,grill2020bootstrap,caron2021emerging,assran2023ijepa}, where model $f$ learns by optimizing \emph{contrastive objectives}. Formally, let $\mathbf{x}', \mathbf{x}''$ be two augmentations of a sample $\mathbf{x} \sim X$. Contrastive objectives enforce the similarity of $f(\mathbf{x}')$ and $f(\mathbf{x}'')$ while avoiding trivial solutions, i.e., producing identical embeddings of unrelated data samples. For this purpose, most joint-embedding methods~\cite{chen2020simple,caron2021emerging} use objective functions requiring large batch sizes.

Due to the large-data requirements and the computational overhead associated with the contrastive objectives, deploying SSL methods in highly distributed Federated environments is challenging in practice~\cite{zhuang2021collaborative,zhuang2022divergence,li2023mocosfl}.

\subsection{Momentum contrastive split federated learning}
\label{sec:mocosfl}

Momentum Contrastive Split Federated Learning (MocoSFL)~\cite{li2023mocosfl} addresses the practical challenges of deploying SSL methods in distributed environments by combining SFL~\cite{thapa2020splitfed} with Momentum Contrastive Learning (MoCo)~\cite{he2020momentum} method.

Similarly to MoCo, the contrastive objective of MocoSFL is calculated with InfoNCE~\cite{oord2018representation} based on the \emph{memory of embeddings} $M$:
\begin{equation}
    \mathcal{L}_{InfoNCE}(\mathbf{z}', \mathbf{z}'', M) = - log \frac{
            exp(\mathbf{z}' \cdot \mathbf{z}'' / \tau)
        }{
            exp(\mathbf{z}' \cdot \mathbf{z}'' / \tau) + \sum_{\mathbf{z}_M \in M} exp(\mathbf{z}' \cdot \mathbf{z}_M / \tau)
        }
    \label{eq:infonce}
\end{equation}
where $\mathbf{z}' = f_\phi(\mathbf{x}')$, $\mathbf{z}'' = f_{EMA(\phi)}(\mathbf{x}'')$, $\mathbf{x}'$ and $\mathbf{x}''$ are two augmentations of a single data sample, and $\mathbf{z}_M$ denotes the embeddings stored in the memory $M$ (after each training step, $M$ is updated with $\mathbf{z}''$ in a first-in-first-out manner). Moreover, $\phi$ are parameters of the \emph{online} model, and $EMA(\phi)$, being the exponential moving average of $\phi$, are parameters of the \emph{momentum} model. Because MocoSFL operates in the SFL setup, each client contains its version of parameters $\phi^c_i$ and $EMA(\phi^c_i)$, and there is one version of parameters $\phi^s$ and $EMA(\phi^s)$ on the server. Similarly, the memory $M$ is also maintained by the server.

To our best knowledge, MocoSFL is the only SSL method operating in a challenging cross-client federated learning with over 100 clients, each contributing as few as 250 data samples from different distributions~\cite{li2023mocosfl}. It is possible by using a large memory of negative examples from all clients, which minimizes the detrimental effect of potentially small batch sizes of individual clients~\cite{bulat2021improving,chen2020simple} and reducing the chance of overfitting to any individual client distribution~\cite{zhang2020federated}. Moreover, relegating most model layers and the contrastive objective to the server reduces the computational burden on the clients~\cite{thapa2020splitfed}.

\section{\our{}: Momentum-Aligned Contrastive Split Federated Learning}

In this section, we first describe the limitations of MocoSFL and show that its performance deteriorates using higher split layers (see Section~\ref{sec:mocosfl_limitations}). Then, we analyze the source of this deterioration and introduce \our{} as a proposed solution (see Section~\ref{sec:our}).

\subsection{Limitations of MocoSFL}
\label{sec:mocosfl_limitations}

The main limitation of MocoSFL is a drop in performance when splitting the network at the higher layers. In consequence, it can be effectively used only with a few layers on the client side. It leads to two crucial concerns. First, sending representations (activations) from low layers makes the model more sensitive to data leakage and attacks such as the Model Inversion Attack (MIA) ~\cite{fredrickson2015mia}. Second, low-layer representations are much larger than those returned by higher layers, which increases communication overhead. Below, we describe those problems in detail.

\begin{figure}[t]
    \centering
    \includegraphics[width=0.75\textwidth]{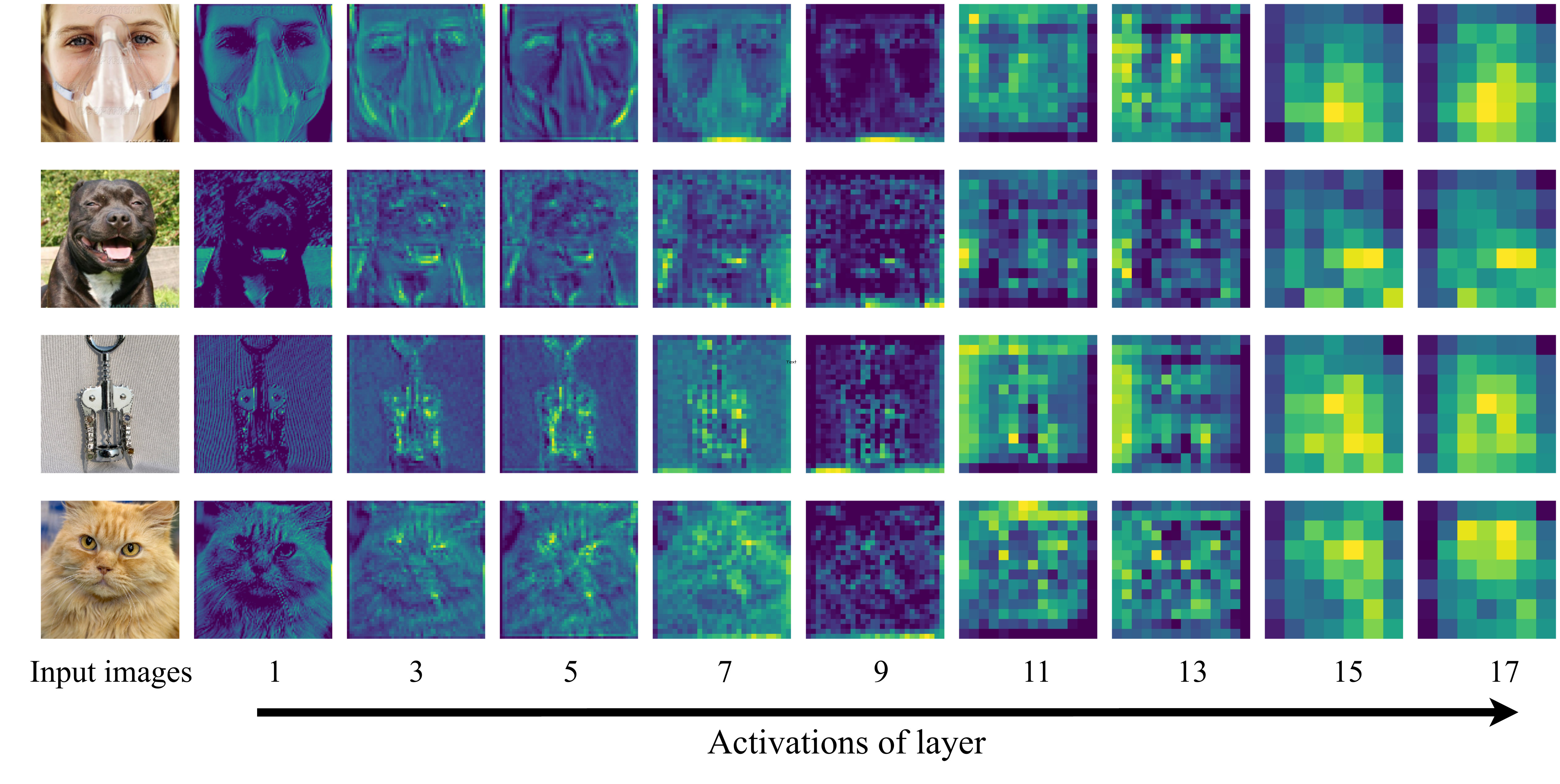}
        \caption{Activations obtained from successive layers of untrained ResNet-18 for a sample ImageNet image. One can observe that representations of low layers highly resemble the respective input data, increasing the privacy risks associated with broadcasting network activations.}
        \label{fig:activations_privacy}
    \vspace{-0.2cm}
\end{figure}

\paragraph{Privacy concerns} caused by sending representations from low layers is illustrated in Figure~\ref{fig:activations_privacy}. One can observe that representations of low ResNet18~\cite{he2015deep} layers highly resemble the respective input data, in contrast to the activations from the higher layers. In fact, in principle, models that learn perceptive features (such as MoCo) do not retain reconstructive features in their high representations~\cite{balestriero2023cookbook}. Thus, in SFL, increasing the number of layers on the client side reduces the privacy risks associated with broadcasting network representations.

\begin{figure}[h]
    \centering
    \includegraphics[width=0.8\textwidth]{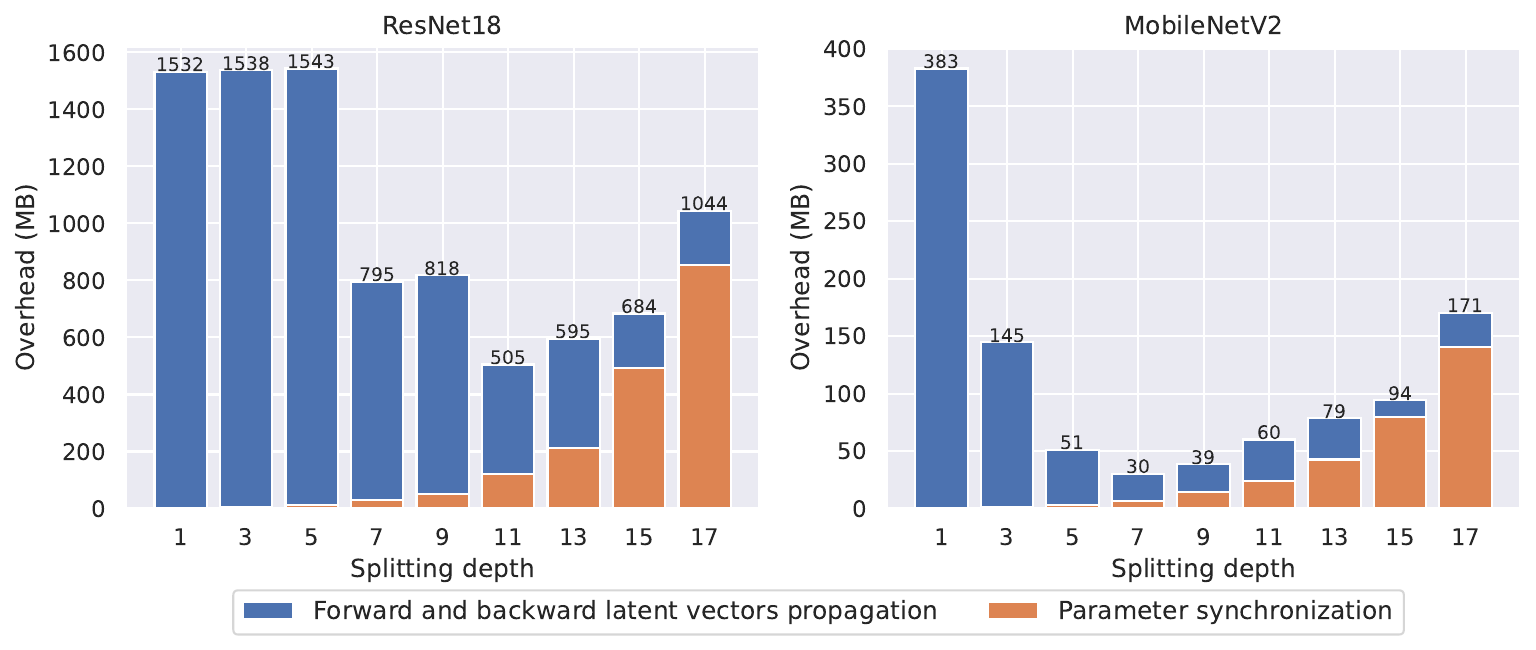}
    \caption{Communication overhead of a single client device for one training epoch of ResNet-18~\cite{he2015deep} and MobileNet-V2~\cite{sandler2018mobilenetv2} for different splitting depths. The 11-th and 7-th layers are the most communication-efficient for ResNet-18 and MobileNetV2, respectively. Note that the training epoch corresponds to $250$ images of resolution $224 \times 224$ are processed, and $10$ synchronizations of parameters. Moreover, the blue bars correspond to communication in the optimization phase, and the orange bars correspond to parameter synchronization.}
    \label{fig:oh_barplots}
\end{figure}

\paragraph{Communication overhead.} Network architectures used in mobile devices, such as ResNet~\cite{he2015deep} or MobileNet~\cite{sandler2018mobilenetv2}, progressively downsample the spacial dimensions of representations obtained from successive layers while simultaneously increasing their channel dimensions. As a result, the overall representation size \emph{decreases} with network depth, and the communication in the SFL optimization phase requires less bandwidth. However, more layers on client devices require exchanging more parameters during the SFL synchronization phase, which translates to a larger bandwidth. Therefore, a trade-off exists between those two communication overheads, as illustrated in Figure~\ref{fig:oh_barplots} with an optimal split in the higher layers.

\begin{figure}[h]
    \centering
    \includegraphics[width=0.65\textwidth]{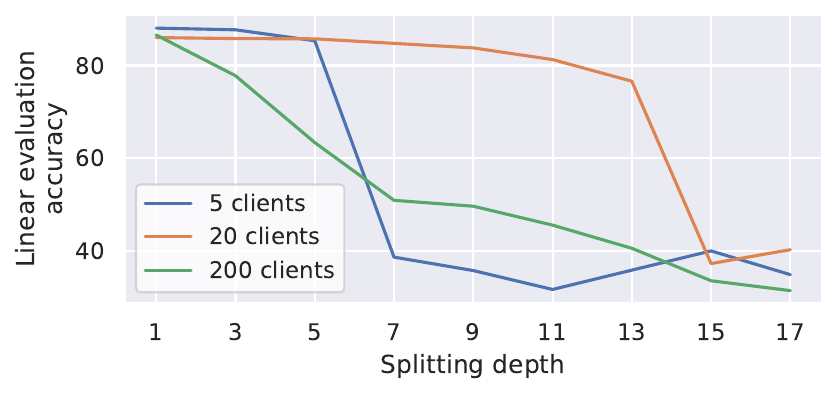}
    \caption{Accuracy of MocoSFL drops significantly with increased splitting depth regardless of the number of clients. Here, presented for CIFAR-10~\cite{krizhevsky2009learning}.}
    \label{fig:mocosfl_deterioration}
\end{figure}

\paragraph{Deteriorated performance for higher split layer.}
Based on the above considerations, it is evident that choosing a larger splitting depth can be attractive due to privacy and efficiency reasons. Thus, a natural question arises -- \textbf{how does splitting depth affect the performance of MocoSFL?} To answer this,
we benchmark its performance on the CIFAR-10 dataset~\cite{krizhevsky2009learning} for 5, 20, and 200 clients, following the experimental setup of~\cite{li2023mocosfl} (see Section \ref{sec:hyperparamters}) and report the results in Figure~\ref{fig:mocosfl_deterioration}
\footnote{Note that Li et. al.~\cite{li2023mocosfl} evaluate MocoSFL only with split layer set to 1 or 3.}. Surprisingly, for higher split layers, we observe a catastrophic deterioration in the accuracy of the trained models.

To summarize, pretraining mobile-friendly architectures with MocoSFL with a splitting depth larger than $7$ is unreliable, and using a lower splitting depth poses concerns to privacy and communication overhead.

\subsection{\our{}}
\label{sec:our}

\begin{figure}[t]
    \vspace{-0.25cm}
    \centering
    \includegraphics[width=\textwidth]{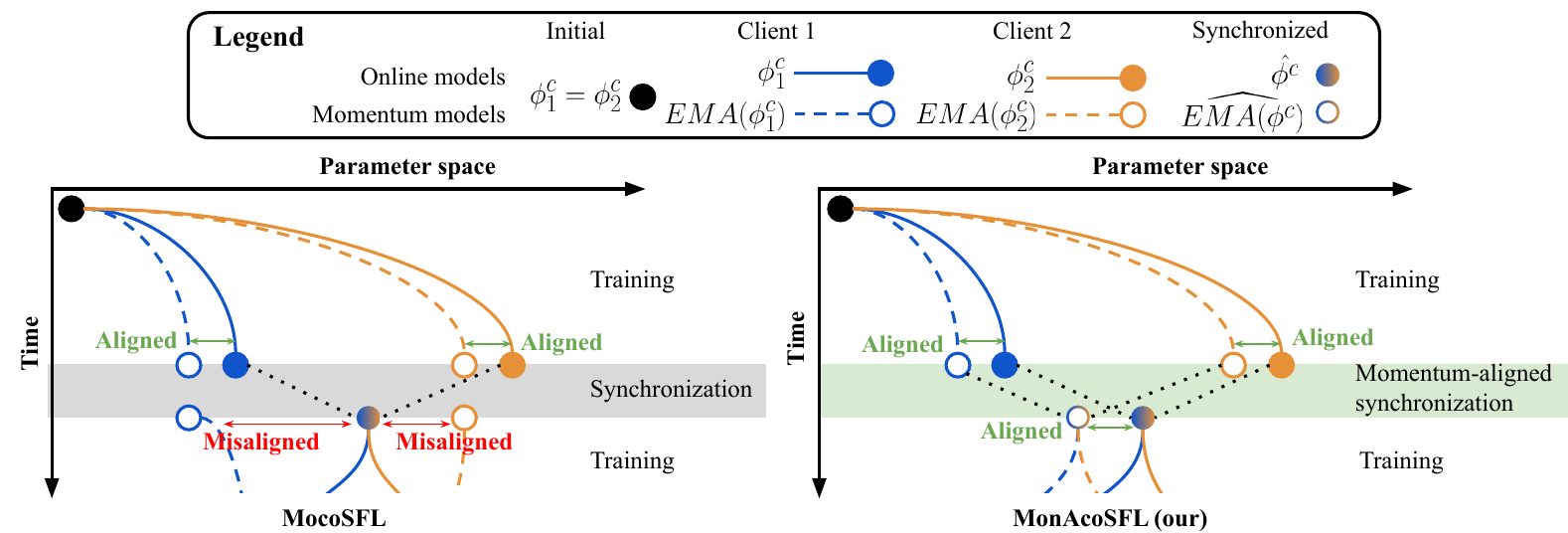}
    \caption{Visualization of parameters changing in MocoSFL (left) and \our{} (right) for two clients. The solid and dashed lines represent the progression of online and momentum parameters, respectively. The dotted lines symbolize the synchronization of parameters. The difference between MocoSFL and \our{} lies in the synchronization procedure that, in the case of \our{}, ensures that both online and the momentum models remain aligned, preserving their ability to optimize the contrastive objective.}
    \label{fig:schema_misalignement}
    \vspace{-0.25cm}
\end{figure}

Before providing our solution to the limitations of MocoSFL, we will first outline the main reason behind them. For this purpose, let us analyze the MocoSFL training procedure in detail. As we described in Section~\ref{sec:ssl_and_federated}, client models start with the same initial parameters $\phi^c_1 = \phi^c_2 = \dots = \phi^c_N$. However, during training, they move away from one another due to different local datasets, and they must be periodically synchronized to stabilize the training. Like in regular SFL schemes, MocoSFL synchronizes \emph{only the online client models}. However, such rapid modification of online parameters breaks the underlying assumption of MoCo that respective online and momentum models encode similar representations at all times, which is essential for minimizing the contrastive objective~\cite{he2020momentum}. We illustrate this phenomenon on the left side of Figure~\ref{fig:schema_misalignement}. This problem grows with the increased splitting depth because more parameters become misaligned, which explains the decrease in performance.

To overcome this limitation, we propose \textbf{Mo}me\textbf{n}tum-\textbf{A}ligned \textbf{Co}ntrastive \textbf{SFL} (\our{}) presented on the right side of Figure~\ref{fig:schema_misalignement}. It ensures the alignment of respective online and momentum client models by synchronizing the respective client momentum models whenever the online models are synchronized, i.e. by overwriting momentum model of each client with:
$$
\widehat{EMA(\phi^c)} = \dfrac{\sum_{i=0}^N EMA(\phi^c_i)}{N}
$$

Note that since $EMA(\phi^c_1), \dots, EMA(\phi^c_N)$ are the EMAs of the individual  $\phi^c_1, \dots, \phi^c_N$, their average always corresponds to the EMA of the average value of the online parameters ($\hat{\phi^c}$), i.e: $\widehat{EMA(\phi^c)} = EMA(\hat{\phi^c}).$

\section{Experimental investigation}

In this section, we experimentally evaluate \our{} and compare it with MocoSFL, a recent state-of-the-art method in self-supervised federated learning. We evaluate both models using classification accuracy on downstream tasks and verifying the privacy of clients data. Finally, we experimentally analyze the alignment of online and momentum models in MoCo framework. For a fair comparison, we closely follow the experimental setup of Li et.al.~\cite{li2023mocosfl}. 


\subsection{Experimental setup} 
\label{sec:hyperparamters}

\paragraph{Hardware} We simulate the distributed environment 
of MocoSFL and \our{} 
on a single machine that hosts the client and server models and runs all of them on a single NVidia A100 GPU. 
We conduct experiments on mobile-friendly ResNet18~\cite{he2015deep} and MobileNetV2~\cite{sandler2018mobilenetv2} backbones. 

\paragraph{Data} We conduct our experiments on CIFAR-10 and CIFAR-100~\cite{krizhevsky2009learning} datasets. We divide the samples equally between the clients to simulate the situation where each client has access to only a small part of the data. We assume a challenging setting where the data is not Independent and Identically Distributed (non-IID) between the client devices. Namely, we assign for each client images from randomly chosen 2 (for CIFAR-10) or 20 (for CIFAR-100) classes.

\paragraph{Number of clients} We compare MocoSFL and \our{} in both cross-silo (5 or 20 clients) and cross-client (200 clients) settings. To our knowledge, MocoSFL is the only previous self-supervised federated method able to scale to the cross-client setting. We synchronize the models on client devices 10 times per training epoch, which amounts to synchronizing the model every 1000, 250, and 25 encountered images for the 5, 20, and 200-client settings, respectively. We adjust the client-side batch size and client sampling ratio accordingly in order to keep the server-side batch size at around 100. For example, in the 5-client setting, the batch size is 20 and all clients are used at each training epoch, whereas in the 200-client setting, we use a random choice of 100 clients in each epoch and therefore each client has a batch size of 1.

\paragraph{SSL model} We use the MoCo-v2~\cite{chen2020improved}, where the backbone is succeeded by a 2-layer MLP  projector network with a hidden size of 1024~\cite{chen2020simple,bordes2022guillotine}, discarded after SSL pretraining. The server maintains a memory $M$ of 6000 negative embeddings, implemented as a First-In-First-Out (FIFO) queue. After each training step, the queue is updated with the momentum model embeddings of the most recent mini-batch of samples. 
We train MocoSFL for 200 epochs, with the SGD optimizer, with an initial learning rate of 0.06, 0.9 momentum, and 0.0005 weight decay. The learning rate is scheduled throughout the training according to the cosine scheduler.

\paragraph{Evaluation} After each epoch, we perform the k-NN validation of the model using 20\% of the validation dataset. K-NN validation is the standard method of evaluating self-supervised representations during their training~\cite{lee2021improving,li2023mocosfl}. After training, we use the model which achieved the best k-NN performance.
We measure the final performance of the model using the widely used linear evaluation protocol~\cite{chen2020simple,grill2020bootstrap,zhuang2022divergence,lee2021improving,li2023mocosfl}. Namely, we attach random liner layer to the frozen pretrained backbone and train only this layer on the labeled dataset for 100 epochs, with a batch size of 128 and the Adam optimizer~\cite{kingma2017adam} with initial learning rate of 0.001 and cosine learning rate scheduler.


\subsection{Accuracy performance}

\begin{figure}[t]
    \centering
    \includegraphics[width=\textwidth]{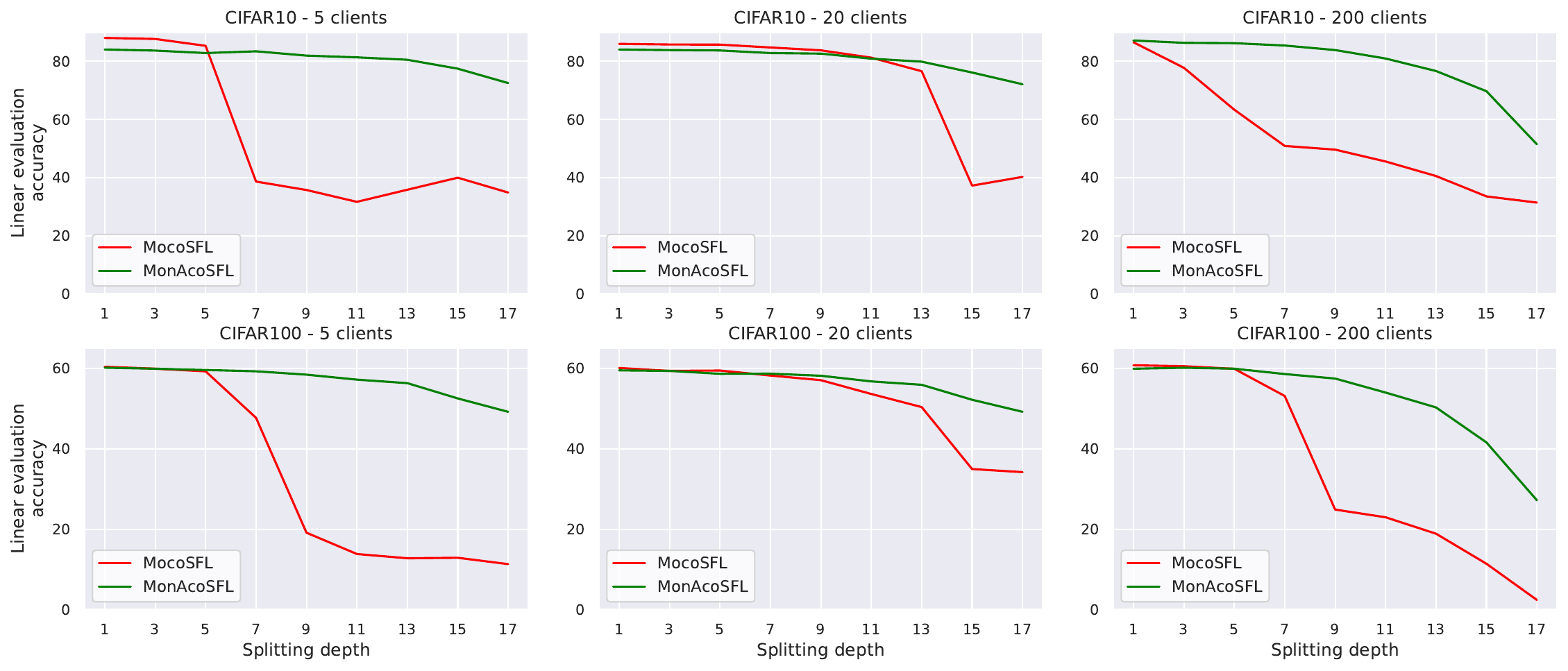}
    \caption{Linear evaluation of MocoSFL and \our{} on ResNet18 architecture.
    \our{} maintains the accuracy with increasing cut-layers, whereas the performance of MocoSFL rapidly deteriorates.}
    \label{fig:accuracy_r18}
\end{figure}

\begin{figure}
    \centering
        \centering
        \includegraphics[width=\textwidth]{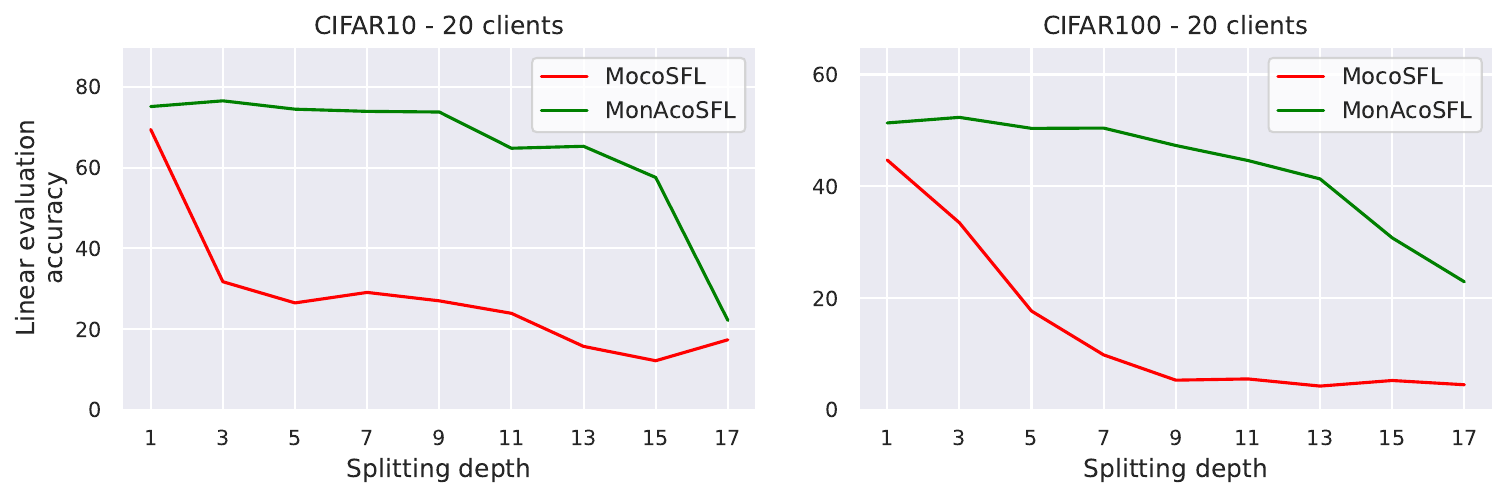}
        \caption{Linear evaluation accuracy on MobileNetV2 backbone trained with MocoSFL and \our{}. There is significant discrepancy between \our{} and MoCoSFL at every splitting depth.}

        
    \label{fig:accuracy_mn}
    \vspace{-0.5cm}
\end{figure}

Figure~\ref{fig:accuracy_r18} presents the accuracy of \our{} and MocoSFL in the cross-silo (5 or 20 clients) and cross-client (200 clients) settings on the ResNet architecture~\cite{krizhevsky2009learning}. Although both models decrease the accuracy with increasing splitting depth, this decrease is much smaller in \our{} than in MocoSFL. 
In the most communication-efficient splitting point (layer 11-13), the difference between \our{} and MoCoSFL exceeds 30 percentage points of accuracy. An even stronger advantage of \our{} is confirmed on the MobileNet backbone, see Figure \ref{fig:accuracy_mn}. In the case of 20 clients, \our{} improves the accuracy of MoCoSFL by more than 40 percentage points starting from 3th to 15th cut layer on the CIFAR-10 dataset.



\subsection{Privacy Evaluation}

We next compare the privacy-preserving capabilities of \our{} and MoCoSFL with a Model Inversion Attack (MIA)~\cite{fredrickson2015mia}. We assume that the attacker has access to 1\% of the training data and trains a decoder for reconstructing images from client model embeddings. 
We train the decoder with the MSE loss and Adam optimization with a learning rate of 0.001 over 50 epochs and a batch size of 32. We perform an attack on ResNet-18 networks pretrained on CIFAR-100 dataset with 20 clients and varying cut-layers. The attack is conducted without pre-training or employing additional privacy-enhancing techniques like TAResSFL~\cite{li2023mocosfl}. 

\begin{figure}[h]
    \centering
    \begin{subfigure}[t]{0.48\textwidth}
        \centering
        \includegraphics[width=\textwidth]{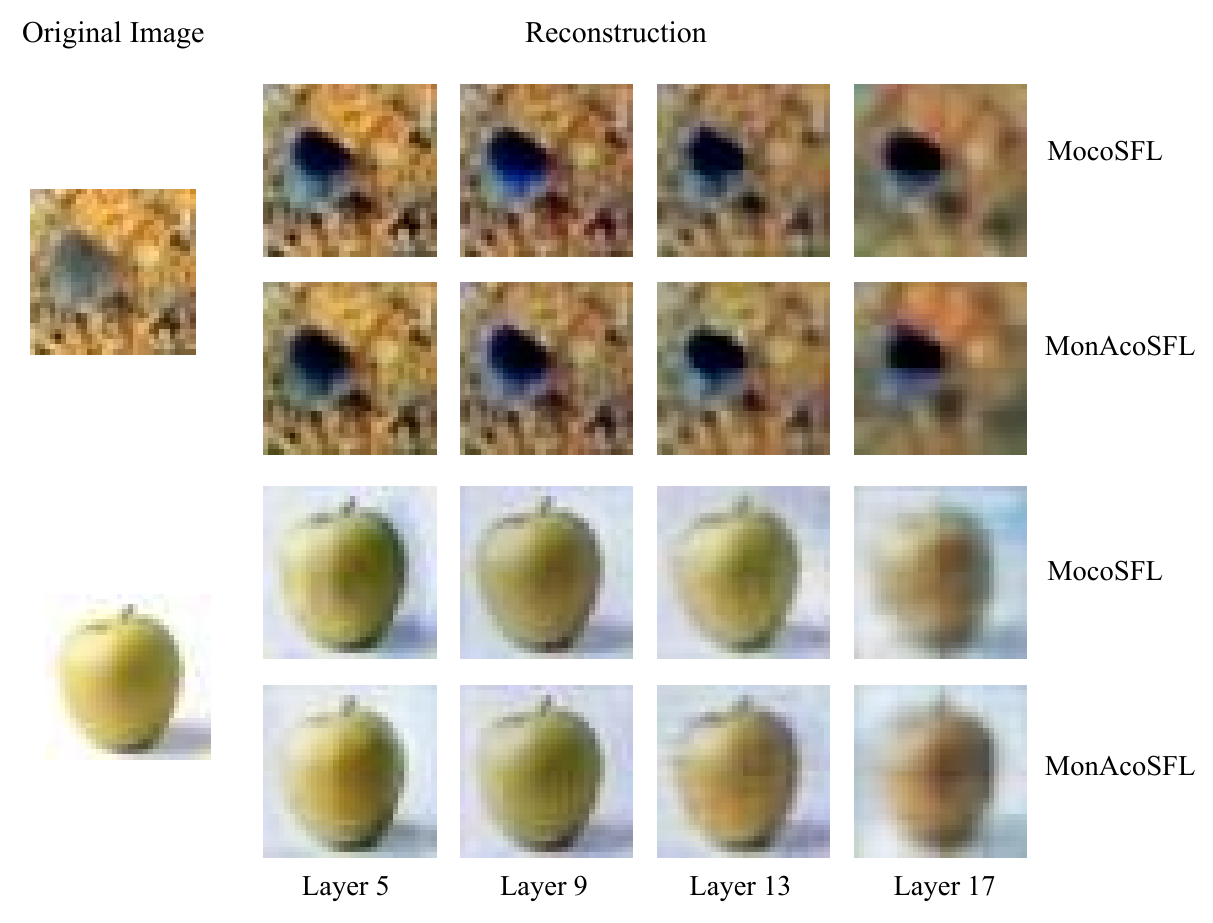}
        \caption{Images reconstructed by the attacker at different layers following MIA on both MocoSFL and \our{}. }
        \label{fig:attack_vis}
    \end{subfigure}
    \begin{subfigure}[t]{0.48\textwidth}
        \centering
        \includegraphics[width=\textwidth]{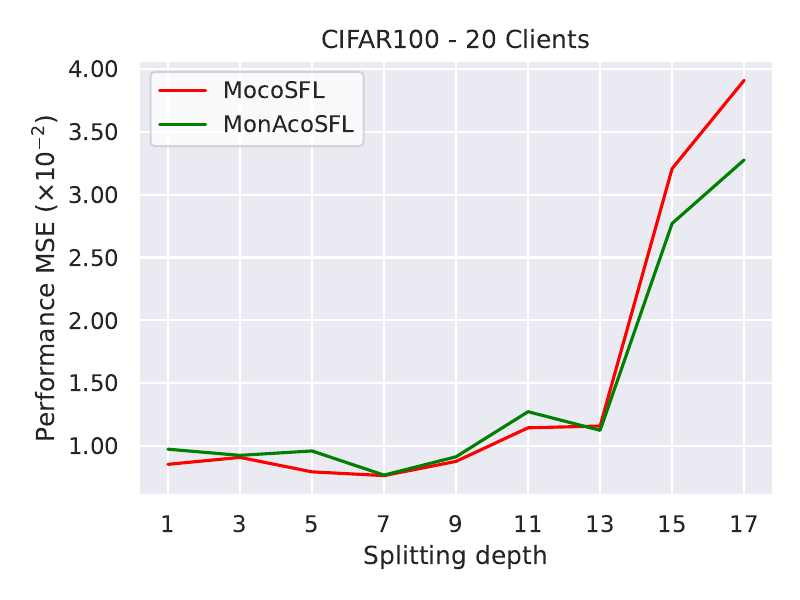}
        \caption{MSE of reconstructing the original images by the attacker. 
        Higher MSE values for deeper cut-layers indicates a better resistance to attack (better privacy).
        }
        \label{fig:mse_r18}
    \end{subfigure}
    \caption{MIA attack on ResNet-18 models trained with MocoSFL and \our{}. Deeper splits provide better protection of client data. Both methods achieve similar levels of privacy protection.}
\end{figure}

We compare the original and reconstructed images in terms of MSE in Figure~\ref{fig:mse_r18}, where a higher MSE indicates better privacy protection. We find that for cut-layers 1-13 the attacks on MocoSFL and \our{} yield similar reconstruction errors, with the errors of attacks on \our{} being larger by a small margin. The MSE remains relatively stable for layers 1-7 and increases for layers 9-17, indicating greater privacy of embeddings of these cut-layers. In layers 15-17, MocoSFL achieves larger attack MSE than \our{}. We attribute this to the fact that the representation of MocoSFL trained with these cut-layers is in general of much worse quality (which results in lower accuracy, see Figure \ref{fig:accuracy_r18}) and, as a side-effect, encodes less information about the data. Concluding, using deeper cut-layers (9-13) is not only more efficient from the computational point of view but also increases the protection of client data.


\subsection{Analysis of models alignment}

To empirically verify that \our{} indeed preserves the alignment of online and momentum model parameters, we measure it throughout the training. We define the average misalignment of online and momentum parameters as their average absolute difference i.e.:
\begin{equation}
    \sum_{i=0}^N \dfrac{|\phi^c_i - EMA(\phi^c_i)|}{N \cdot dim(\phi^c)},
\end{equation}
where $N$ denotes the number of clients and $dim(\phi^c)$ is the dimension of client model parameters.

\begin{figure}[h]
    \centering
        
        \centering
        \includegraphics[width=0.7\textwidth]{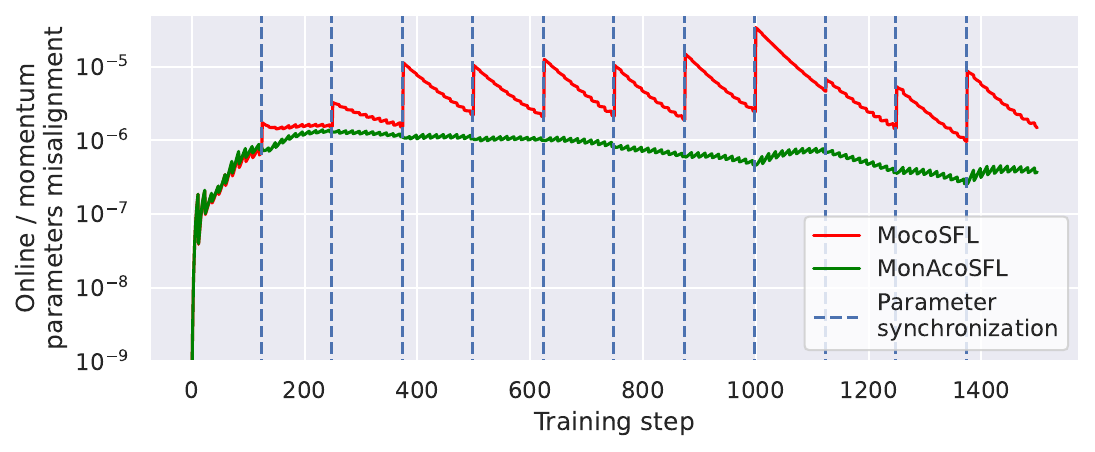}
        \caption{A comparison of average misalignment of online and momentum client models in the initial steps of training for \our{} and MocoSFL (lower is better). We mark with blue lines the steps at which the parameter synchronization is performed (every 125 steps). In MocoSFL the misalignment grows rapidly during parameter synchronizations, whereas in \our{} it remains an order of magnitude smaller, resulting in a more stable training. }
        \label{fig:misalignment_plot}
        \vspace{-0.5cm}

        
\end{figure}

We plot in Figure~\ref{fig:misalignment_plot} the misalignment values for 1500 initial training steps (out of approximately 50000) of ResNet18 trained on CIFAR-100 by 20 clients, MocoSFL and \our{}, with split at the 11th layer. Throughout the first 125 steps, both methods display similar misalignments. However, during parameter synchronizations, the difference between the online and momentum models in MocoSFL rapidly increases by an order of magnitude. On the other hand, in \our{}, the misalignment of momentum and online models does not change significantly throughout the training.

\section{Conclusion}

In this paper, we conducted an in-depth analysis of MocoSFL -- the state-of-the-art method for Federated Self-Supervised Learning~\cite{li2023mocosfl}. We found that the quality of this method deteriorates in privacy-preserving and communication-efficient settings and identified that the reason for this is its synchronization scheme, which leaves the participating online and momentum models misaligned. 
We addressed this problem by introducing \textbf{Mo}me\textbf{n}tum-\textbf{A}ligned \textbf{co}ntrastive \textbf{S}plit \textbf{F}ederated \textbf{L}earning (\our{}) and showed that it significantly improves the performance of MoCoSFL, especially under communication-efficient conditions. 

From a theoretical perspective, our solution is centered on the correct synchronization of online and momentum models of self-supervised learners deployed in distributed environments. Given that the online/momentum model pairs are a staple of several modern SSL approaches~\cite{grill2020bootstrap,caron2021emerging,assran2023ijepa}, our findings can be applicable in federated learning of various SSL methods other than MoCo~\cite{he2020momentum}.
From the practical perspective, \our{} is the first Federated self-supervised method to achieve good results using deeper cut layers, providing efficient communication between clients and server and better protection of client data. 

\section*{Ethical implications}
In this work, we focus on improving the existing Federated self-supervised approaches in regimes where data privacy and communication efficiency are of concern. Our findings can lead to reducing the risks of data leakages when deploying Federated self-supervised learning algorithms in real-world scenarios.

\section*{Acknowledgements}
This research has been supported by the flagship project entitled "Artificial Intelligence Computing Center Core Facility" from the Priority Research Area DigiWorld under the Strategic Programme Excellence Initiative at Jagiellonian University, and by the Horizon Europe Programme (HORIZON-CL4-2022-HUMAN-02) under the project "ELIAS: European Lighthouse of AI for Sustainability", GA no. 101120237.
The research of Marcin Przewięźlikowski was supported by the National Science Centre (Poland), grant no. 2023/49/N/ST6/03268.
The research of Marcin Osial was supported by National Science Centre (Poland) grant number 2023/50/E/ST6/00469.
The research of Marek Śmieja was supported by the National Science Centre (Poland), grant no. 2022/45/B/ST6/01117. 
The research of Bartosz Zieliński was supported by National Science Centre (Poland) grant number 2022/47/B/ST6/03397.
Some experiments were performed on servers purchased with funds from a grant from the Priority Research Area (Artificial Intelligence Computing Center Core Facility) under the Strategic Programme Excellence Initiative at Jagiellonian University.
We gratefully acknowledge Polish high-performance computing infrastructure PLGrid (HPC Center: ACK Cyfronet AGH) for providing computer facilities and support within computational grant no. PLG/2023/016303.
For the purpose of Open Access, the author has applied a CC-BY public
copyright license to any Author Accepted Manuscript (AAM) version arising
from this submission.



%
\bibliographystyle{splncs04}
\bibliography{ref}

\end{document}